# Automatic question generation for propositional logical equivalences

Yicheng Yang
CST
BNU-HKBU United International College
Zhuhai Guangdong China
yangyicheng002@gmail.com

Xinyu Wang
CST
BNU-HKBU United International College
Zhuhai Guangdong China
amberwangxinyu@icloud.com

Haoming Yu
CST
BNU-HKBU United International College
Zhuhai Guangdong China
yu2974201195@gmail.com

Zhiyuan Li
CST
BNU-HKBU United International College
Zhuhai Guangdong China
goliathli@uic.edu.cn

## ABSTRACT

The increase in academic dishonesty cases among college students has raised concern, particularly due to the shift towards online learning caused by the pandemic. We aim to develop and implement a method capable of generating tailored questions for each student. The use of Automatic Question Generation (AQG) is a possible solution.

Previous studies have investigated AQG frameworks in education, which include validity, user-defined difficulty, and personalized problem generation. Our new AQG approach produces logical equivalence problems for Discrete Mathematics, which is a core course for year-one computer science students. This approach utilizes a syntactic grammar and a semantic attribute system through top-down parsing and syntax tree transformations.

Our experiments show that the difficulty level of questions generated by our AQG approach is similar to the questions presented to students in the textbook [1]. These results confirm the practicality of our AQG approach for automated question generation in education, with the potential to significantly enhance learning experiences.

## CCS CONCEPTS

• Mathematics of computing~Discrete mathematics • Theory of computation~Logic~Equational logic and rewriting • Theory of computation~Formal languages and automata theory~Grammars and context-free languages

## 1 Introduction

In recent years, the growing usage of online resources in academia has raised concerns about academic dishonesty. According to a report from China Education Daily, the number of plagiarism cases among college students has greatly increased since the outbreak of COVID-19 [2]. The report has indicated that the shift to online learning, combined with the study pressure, has made cheating easier. In another survey conducted by China Youth Daily, 71.5% of college students admitted to witnessing or participating in academic misconduct, including plagiarism, during the pandemic [3]. This is a significant increase from the 52.3% reported by the same survey conducted before the pandemic. The survey has also found that 38.1% of students believe that online learning has increased the easiness of academic misconduct. Another paper [4] has reported the increasing of plagiarism risk in Taiwan as the growth of online learning resources. One reason for the increase of plagiarism cases is the lack of invigilation on assessments. According to Xinhua News Agency, "the absence of the physical learning environment and the inability of teachers to monitor students' behavior in real time create a favorable environment for academic misconduct." [5] The above data shows that online classes and the epidemic have a significant impact on college students' plagiarism tendencies. As educators in Computer Science, we have zero tolerance for cheating.

The most effective way to minimize plagiarism is giving different questions to different students. If any two students are given totally different questions, they can no longer copy from each other.

There are several ways to implement it. One commonly used trick is to create a manually designed question bank. When assignments or exam papers are created, we use a hash function to hash students to different questions. However, this method has some limitations if the student population is large. To minimize the hash collision, the question bank has to be large, which is a heavy workload for the designer. Besides, if the question bank is not regularly updated, some questions can possibly be leaked to students.

Another common trick is parameterizing questions using a piece of students' personal information, like student ID. For example, the original question is "Find the root of the quadratic function $f(x) = x^2 - 1$" which can be parameterized as "Find the root of the quadratic function $f(x) = x^2 - a$, where $a$ is the last digit of your student ID". Nevertheless, this method only applies to calculation questions for courses like Calculus or Linear Algebra. But for proof problems in Discrete Mathematics, a fundamental math course for year one computer science students, parameterization does not work well. Proofs, different from calculations, do not have any constant to be parameterized. For example, Prove that "$p \equiv p \lor (p \land q)$". In this equivalence, $p$ and $q$ are propositional variables.

Thus, this research intends to design a question generator for Discrete Math. Our generator ensures:
- different students get different questions;
- generated questions are of the same level of difficulty; and
- students cannot get new questions by regeneration.

For the first step of the big picture, we target the questions on logical equivalences. Students are given two expressions in propositional logic and required to show that the expressions are equivalent using some laws. We choose this type of question because they appear in the first chapter of Discrete Mathematics.

Automatic Question Generation is an existing class of algorithms to generate questions, which has been studied by many scholars. In general, there are 3 stages involved in the AQG algorithm, a brief process description is as follows.

1. Preprocessing: If the input is a question specification in natural language, preprocessing interprets the input using some natural language processing methods.
2. Question construction: This is the main stage of the generation, which constructs the stem of a question and decoys (for example, incorrect options for multiple choices)
3. Post-processing: This stage creates a context that fits the question stem.

Kurdi et al. [6] presented a systematic review of the results of research on the problem of automated generation for educational purposes. The results showed that the study of AQG has gained wide attention and research in recent years. Singhal et al. [7] describe a framework which could help teachers quickly generate large numbers of questions on a geometry topic for high school mathematics. This method ensures the validity of the generated questions. Singhal and Henz [8] also introduced a framework to limit the scope of knowledge the question examines. The proposed method initially obtains input from the user. Mathematically, the input for generating geometry question can be represented by a tuple $(S, Q)$ where:
- $S$ is a multiset of shapes, such as "line", "triangle", etc.
- $Q \subseteq$ {"area", "perimeter", "length", "angle"}

Based on this input, a *region graph* is generated which represents the regions formed by the object. Similarly, the teacher can generate the required questions in our framework by entering limitations such as grammar rules or question length.

Singhal et al. proposed a framework [9], which considers user-defined difficulty for AQG. It contains a user-priority list which allows users to predefine the difficulty factors, such as the number of blocks, number of pulleys, acceleration of blocks, etc. Consequently, the difficulty level of a question is counted as the sum of all the factors.

Several other papers present a range of relevant frameworks as well. Polozov et al. [10] introduced how to make personalized problems based on individual students' characteristics and preferences. The system utilizes machine learning techniques and natural language processing to create personalized math exercises, tailoring the difficulty level, context, and topics to suit each student's learning needs. Wang [11] and Gupta [12] proposed some techniques to generate mathematical problems. Both of them utilize natural language processing. Gupta uses a problem-based approach to enhance students' understanding of mathematical concepts and create an interactive and adaptive learning experience, while Wang focuses on mathematical reasoning to create a diverse set of math problems that require different problem-solving strategies. Zavala [13] and Thomas et al. [14] provided a good tool to generate programming exercises. Zavala focuses on using Semantic-Based Abstract Interpretation Guidance (AIG) to automatically generate programming exercises with varying levels of difficulty while Thomas presents a stochastic tree-based method for generating program-tracing practice questions that require learners to understand code execution flow. For more about AQG for computer science or mathematics, readers are referred to [15] [16] [17].

Unfortunately, none of the existing methods fulfills our requirements listed above. We have also tried GPT, the popular large language model. However, GPT generation is not robust. It sometimes produces two propositions which are not equivalent. Thus, this paper proposes our new method, based on the syntax and semantics of logical equivalences.

Section 2 provides preliminary information that is crucial for comprehending the entirety of our research. Section 3 proceeds to elaborate on the implementation process and the specific methodologies employed in this study. Section 4 presents experiment results to compare the performance of our generator with human-designed questions. Section 5 concludes this paper.

## 2   Preliminaries

A logical equivalence question gives two *propositions*. Students need to show that the two propositions are *logically equivalent* using *equivalence laws*. A proposition can be *atomic* or *compound*. An atomic proposition is either a constant ($True$ or $False$) or a propositional variable (English letters in lower case). A compound proposition can be a *unary operator* followed by a proposition, or two propositions connected by a *binary operator*. In this paper, we only discuss the unary operator *negation* $\neg$, and the binary operators *conjunction* $\land$, *disjunction* $\lor$, *implication* $\subset$, and *biconditional* ↔. Please note that we use entailment ⊂ to denote implication in this paper because right arrow → is captured by grammar productions. The algebraic definitions of these operators are in Table 1 and the *precedence* is in Table 2 *Precedence of logical operators* (a smaller value represents a higher precedence).

| $P$ | $Q$ | $\neg P$ | $P \wedge Q$ | $P \vee Q$ | $P \subset Q$ | $P \leftrightarrow Q$ |
|---|---|---|---|---|---|---|
| T | T | F | T | T | T | T |
| T | F | F | F | T | F | F |
| F | T | T | F | T | T | F |
| F | F | T | F | F | T | T |

Table 1 *Definition of logical operators*

| Operator | Precedence |
|---|---|
| ¬ | 1 |
| ∧ | 2 |
| ∨ | 3 |
| ⊂ | 4 |
| ↔ | 5 |

Table 2 *Precedence of logical operators*

Two propositions $(P, Q)$ are *equivalent*, denoted as $P \equiv Q$, if $P$ and $Q$ have the same truth value for any possible assignment of propositional variables. Other than drawing a truth table, one can also prove equivalences using the following equivalence laws.

| | | |
|---|---|---|
| ***Identity*** | $p \wedge T \equiv p$ | $p \vee F \equiv p$ |
| ***Domination*** | $p \wedge F \equiv F$ | $p \vee T \equiv T$ |
| ***Commutative*** | $p \wedge q \equiv q \wedge p$ | $p \vee q \equiv q \vee p$ |
| ***Idempotent*** | $p \vee p \equiv p$ | $p \wedge p \equiv p$ |
| ***Negation*** | $p \wedge \neg p \equiv F$ | $p \vee \neg p \equiv T$ |
| ***Absorption*** | $p \vee (p \wedge q) \equiv p$ | $p \wedge (p \vee q) \equiv p$ |
| ***Associative*** | $(p \wedge q) \wedge r \equiv p \wedge (q \wedge r)$ | |
| | $(p \vee q) \vee r \equiv p \vee (q \vee r)$ | |
| ***De Morgan*** | $\neg (p \wedge q) \equiv \neg p \vee \neg q$ | |
| | $\neg (p \vee q) \equiv \neg p \wedge \neg q$ | |
| ***Double Negation*** | $\neg\neg p \equiv p$ | |
| ***Distributive*** | $p \vee (q \wedge r) \equiv (p \vee q) \wedge (p \vee r)$ | |
| | $p \wedge (q \vee r) \equiv (p \wedge q) \vee (p \wedge r)$ | |

The syntax of a context-free language is defined as a *context-free grammar* (CFG). In general, a CFG is a 4 tuple $(\mathcal{N}, \mathcal{T}, \mathcal{R}, S)$, where

- $\mathcal{N}$ is a finite set called non-terminals,
- $\mathcal{T}$ is a finite set, disjoint from $N$, called the terminals,
- $\mathcal{R}$ is a finite set of *production rules*, with each rule being a non-terminal $V$ and a string $a_1 a_2 \cdots a_n$ of non-terminals and terminals in the form $V \to a_1 a_2 \cdots a_n$, and
- $S \in \mathcal{N}$ is the start non-terminal.

We denote $V \to a|b$ by the abbreviation of $V \to a$ and $V \to b$. Let $V \to a_1 a_2 \cdots a_n$ be a production rule and "…$V$…" be a string which contains $V$. A *substitution* is replacing the existence of $V$ in the string, and the resulting string is "… $a_1 a_2 \cdots a_n$ …", which is denoted by …$V$…⇒…$a_1 a_2 \cdots a_n$…. In addition, a *production* is a sequence of substitutions, which starts from the start non-terminal $S$ and ends by a string of terminals. Moreover, the entire process of the generation can be represented by a *syntax tree*, where non-terminals are internal vertices, terminals are leaves, and vertex $V$ has children $a_1, a_2, \cdots, a_n$ if the rule $V \to a_1 a_2 \cdots a_n$ is used in one step of a production.

An *attribute system* of a formal language consists of a set of *attribute definitions* for all non-terminals and terminals. The attribute of a terminal depends on the terminal itself, which call *intrinsic*. The attribute definition of a non-terminal $V$ is associated with the production rules in which $V$ is involved. Assume $V \to a_1 a_2 \cdots a_n$ is a production rule, the attribute definition $V.at \coloneqq f(a_1.at, a_2.at, \cdots a_n.at)$ means that the attribute of $V$ depends on the attributes of $a_1 \cdots a_n$, which is called *synthesized*. If the attribute definition is $a_1.at \coloneqq f(V.at, a_2.at, \cdots, a_n.at)$, the attribute is *inherited*. If all attributes of terminals and non-terminals can be computed by a single DFS on the syntax tree, the language is *L-attributed*. In compilers, attribute definitions are used to check types of non-terminals, evaluate computations, and construct target expressions.

Readers are referred to [1] for more details of propositional logic and logical equivalences, and [18] for more about syntax and semantics of a formal language.

## 3  Method

Our algorithm takes a piece of students' information and a set of hyperparameters from instructors as input, produces two equivalent propositions, and consists of 3 major phases.

1. Random seed generation
2. Syntax tree generation
3. Propositional expression construction

The first phase converts students' information into MD5 codes as random seeds, which guarantees a low probability of hash collision [19]. Furthermore, using random seeds allows our algorithm generates the same set of problems for a student multiple times. Then, students cannot get easier questions by smashing the "generate" button.

The second phase generates a syntax tree from top to bottom. Each step of the generation takes one digit from the MD5 code and construct a substructure of the expression. This phase also works with a difficulty control, which uses the hyperparameters from instructors to decide the length of propositions and applications of equivalence laws in the question.

The last phase implements an attribute system and constructs two equivalent propositions from the syntax tree. Our attribute system is carefully designed to be L-attributed, which can be efficiently computed in linear time.

### 3.1 Syntax tree generation

The most natural generation generates one of the two expressions via a syntactical grammar and subsequently applies some equivalence laws on the expression randomly to obtain the second expression. However, this method is not tractable due to the computational complexity. To ensure equivalence laws are correctly applied, the generation needs an attribute system to check the pattern of subexpressions. For example, the absorption rule can be applied to $(E_1 \vee p) \wedge E_2$ only if the subexpressions $E_1$ and $E_2$ are identical. Considering there are multiple equivalence laws can possibly be applied to each subexpression, the computational complexity goes up to exponential.

Thus, a strategic alteration is introduced into the syntax structure by incorporating the symbol $\equiv$ as a terminal in our grammar. In tandem with this modification, the two resulting expressions are connected by $\equiv$ and our generation constructs the structure of two expressions simultaneously. The set of production rules $\mathcal{R}$ is partitioned into two subsets $\mathcal{R}_1$ and $\mathcal{R}_2$. Each rule of $\mathcal{R}_1$ is in the form

$$\alpha_1 E_1 \beta_1 \equiv \alpha_2 E'_1 \beta_2 \rightarrow \alpha_1 \gamma \beta_1 \equiv \alpha_2 \gamma \beta_2$$

for constructing the syntactic structure, where $\alpha_1, \alpha_2, \beta_1, \beta_2$, and $\gamma$ are arbitrary strings of non-terminals and terminals. By applying this substitution, our generation ensures that the subexpression obtained from $E_1$ and $E_2$ are the same.

Each rule of $\mathcal{R}_2$ is in the form

$$\alpha_1 E_1 \beta_1 \equiv \alpha_2 E_1 \beta_2 \rightarrow \alpha_1 \gamma_1 \beta_1 \equiv \alpha_2 \gamma_2 \beta_2$$

to apply the equivalence law $\gamma_1 \equiv \gamma_2$.

Note that this grammar is context-sensitive but not context-free because the left-hand side of each production rule is not a single non-terminal.

Formally, our grammar is as follows.

$\mathcal{N} = \{I\} \cup \{E_i \mid i \geq 0\} \cup \{E'_i \mid i \geq 0\}$
$\mathcal{T} = \{\equiv, \neg, \wedge, \vee, \subset, literal, T, F\}$
$\mathcal{R} = \mathcal{R}_1 \cup \mathcal{R}_2$, where
$\mathcal{R}_1 = \{ I \rightarrow E_0 \equiv E'_0, E_i \rightarrow literal,$
$\quad \alpha_i E_i \beta_i \equiv \alpha'_i E'_i \beta'_i$
$\quad \rightarrow \alpha_i E_j \wedge E_k \beta_i \equiv \alpha'_i E_j \wedge E_k \beta'_i$
$\quad | \ \alpha_i E_j \vee E_k \beta_i \equiv \alpha'_i E_j \vee E_k \beta'_i$
$\quad | \ \alpha_i E_j \subset E_k \beta_i \equiv \alpha'_i E_j \subset E_k \beta'_i$
$\quad | \ \alpha_i \neg E_j \beta_i \equiv \alpha'_i \neg E_j \beta'_i \ \}$,
$\mathcal{R}_2 = \{$
$\quad \alpha_i E_i \beta_i \equiv \alpha'_i E'_i \beta'_i$
$\quad \rightarrow \alpha_i E_j \wedge E_k \beta_i \equiv \alpha'_i E_k \wedge E_j \beta'_i$
$\quad | \ \alpha_i E_j \wedge T \beta_i \equiv \alpha'_i E_j \beta'_i$
$\quad | \ \alpha_i E_j \vee F \beta_i \equiv \alpha'_i E_j \beta'_i$
$\quad | \ \alpha_i E_j \wedge F \beta_i \equiv \alpha'_i F \beta'_i$
$\quad | \ \alpha_i E_j \vee T \beta_i \equiv \alpha'_i T \beta'_i$
$\quad | \ \alpha_i E_j \wedge \neg E_j \beta_i \equiv \alpha'_i F \beta'_i$
$\quad | \ \alpha_i E_j \vee \neg E_j \beta_i \equiv \alpha'_i T \beta'_i$
$\quad | \ \alpha_i E_j \vee E_k \beta_i \equiv \alpha'_i E_k \vee E_j \beta'_i$
$\quad | \ \alpha_i E_j \wedge E_j \beta_i \equiv \alpha'_i E_j \beta'_i$
$\quad | \ \alpha_i E_j \vee E_j \beta_i \equiv \alpha'_i E_j \beta'_i$
$\quad | \ \alpha_i E_j \vee (E_j \wedge E_k) \beta_i \equiv \alpha'_i E_j \beta'_i$
$\quad | \ \alpha_i E_j \wedge (E_j \vee E_k) \beta_i \equiv \alpha'_i E_j \beta'_i$
$\quad | \ \alpha_i E_j \wedge E_k \wedge E_l \beta_i \equiv \alpha'_i E_k \wedge E_j \wedge E_l \beta'_i$
$\quad | \ \alpha_i E_j \vee E_k \vee E_l \beta_i \equiv \alpha'_i E_k \vee E_j \vee E_l \beta'_i$
$\quad | \ \alpha_i \neg (E_j \wedge E_k) \beta_i \equiv \alpha'_i \neg E_j \vee \neg E_k \beta'_i$
$\quad | \ \alpha_i \neg (E_j \vee E_k) \beta_i \equiv \alpha'_i \neg E_j \wedge \neg E_k \beta'_i$
$\quad | \ \alpha_i \neg \neg E_j \beta_i \equiv \alpha'_i E_j \beta'_i$
$\quad | \ \alpha_i E_j \wedge (E_k \vee E_l) \beta_i \equiv \alpha'_i E_j \wedge E_k \vee E_j \wedge E_l \beta'_i$
$\quad | \ \alpha_i E_j \vee (E_k \wedge E_l) \beta_i \equiv \alpha'_i (E_j \vee E_k) \wedge (E_j \vee E_l) \beta'_i \}$,
$S = I$

Notable, $\mathcal{R}_2$ contains 19 rules and looks tedious because we need one production rule for each equivalence law. Some of these production rules have parentheses, but parentheses are not terminals or non-terminals in the grammar. They are only for presenting the precedence of operators and will be directly omitted by the syntax tree.

With the above grammar definition, the syntax tree generation is presented in Algorithm 1. This is a recursion algorithm, which is initially called on the expression $E_1$ because $I \rightarrow E_1 \equiv E'_1$ is always the first rule in the production. The algorithm takes two hex digits from the MD5 code each time. One digit decides whether the algorithm chooses a rule from $\mathcal{R}_1$ to create a syntactic structure or a rule from $\mathcal{R}_2$ to apply an equivalence law. The second digit decides which rule is selected in each category. If all digits are used, $nextHex$ simply cyclic shifts back to the first digit. Furthermore, we can maximize the randomness by applying an index offset each time when every digit is selected exactly once. The offset needs to be relatively prime to 16. A coefficient $\frac{16}{5}$ is applied to line 6 because 5 rules in $\mathcal{R}_1$ can be used in the production. But we want to compare it with a value from 0 to 15. A similar coefficient is applied to line 12.

**Algorithm 1.** SyntaxTreeGeneration $(C, G, V, o_1, o_2, p)$

```
Input:   C = c₁c₂⋯c₁₆ is an MD5 code;
         G is a grammar defined above, such that
         ℛ₁ and ℛ₂ are the two sets of rules of G;
         V ∈ 𝒩 ∪ 𝒯 is a root of a syntax subtree;
         o₁:ℛ₁ → ℕ is a total ordering on ℛ₁;
         o₂:ℛ₂ → ℕ is a total ordering on ℛ₂;
         p ∈ [0,1) is a probability.
Output:  a syntax tree T
1.    c₁ = nextHex(C)
2.    c₂ = nextHex(C)
3.    if V is a terminal then
4.        does nothing and simply returns
5.    else if c₁ < 16 * p₁ then
6.        Let R ∈ ℛ₁ be the rule such that
          o₁(R) is the largest order
          and 16/5 * o₁(R) > c₂.
7.        Assume V is substituted into a₁⋯aₙ by
          applying the rule R.
8.        Construct a subtree rooted at V with
          children a₁, ⋯, aₙ.
9.        if aᵢ is a non-terminal Eₖ, Eₖ has to be
          a new non-terminal which has not been
          used by this algorithm.
10.       for each aᵢ
11.           SyntaxTreeGeneration(C, G, aᵢ, o₁, o₂, p)
12.   else
13.       Does the similar thing but using the
          rules from ℛ₂ and the order o₂ except
          that the coefficient is 16/19.
14.   endif
```

## 3.2 Logical expression construction

Following the syntax generation, the subsequent phase necessitates semantic analysis, which constructs the final logical expression from the syntax tree. This phase involves the establishment of an attribute system to ascertain the interpretation of each non-terminal's expression within the syntax tree.

The attribute system defines two attributes for each grammar symbol, "$exp$" - the expression produced by the grammar symbol and "$pre$" the precedence of the last performed operator in the expression, which is for properly parenthesizing of the logical expression.

First, intrinsic attributes are defined for terminals. For operators, taking $\vee$ as an example,

$$\vee.exp \coloneqq \vee \text{ and } \vee.pre \coloneqq 3$$

meaning that the logical expression of the grammar symbol $\vee$ is itself, and its precedence is 3.

For literals,

$$literal.exp \coloneqq Random() \text{ and } literal.pre \coloneqq \infty$$

The expression of a literal is a propositional variable, which is randomly selected from a variable pool. The precedence of a literal is undefined and assigned to be $\infty$ for computation convenience.

For non-terminals, $exp$ and $pre$ are defined for each production rule. To keep the presentation clean and elegant, we decide to take one production from each of $\mathcal{R}_1$ and $\mathcal{R}_2$ for demonstration purpose (see Figure 1 and Figure 2). Other attribute definitions are omitted. But readers should know that those definitions can be easily constructed in the same way.

---

if $(E_j.pre < \wedge.pre \text{ and } E_k.pre < \wedge.pre)$ then
  $E_i.exp \coloneqq (E_j.exp) \wedge (E_k.exp)$
else if $(E_j.pre < \wedge.pre \text{ and } E_k.pre > \wedge.pre)$ then
  $E_i.exp \coloneqq (E_j.exp) \wedge E_k.exp$
else if $(E_j.pre > \wedge.pre \text{ and } E_k.pre < \wedge.pre)$ then
  $E_i.exp \coloneqq E_j.exp \wedge (E_k.exp)$
else
  $E_i.exp \coloneqq E_j.exp \wedge E_k.exp$
$E'_i.exp \coloneqq E_i.exp$
$E_i.pre \coloneqq \wedge.pre$
$E'_i.pre \coloneqq \wedge.pre$

---

Figure 1. Attribute definition for
$\alpha_i E_i \beta_i \equiv \alpha'_i E'_i \beta'_i \rightarrow \alpha_i E_j \wedge E_k \beta_i \equiv \alpha'_i E_j \wedge E_k \beta'_i$

---

if $(E_k.pre < \vee.pre)$ then
  if $(E_j.pre < \vee.pre)$ then
    $E_i.exp \coloneqq (E_j.exp) \wedge ((E_j.exp) \vee (E_k.exp))$
  else if $(E_j.pre < \wedge.pre \text{ and } E_j.pre > \vee.pre)$ then
    $E_i.exp \coloneqq (E_j.exp) \wedge (E_j.exp \vee (E_k.exp))$
  else
    $E_i.exp \coloneqq E_j.exp \wedge (E_j.exp \vee (E_k.exp))$
else
  if $(E_j.pre < \vee.pre)$ then
    $E_i.exp \coloneqq (E_j.exp) \wedge ((E_j.exp) \vee E_k.exp)$
  else if $(E_j.pre < \wedge.pre \text{ and } E_j.pre > \vee.pre)$ then
    $E_i.exp \coloneqq (E_j.exp) \wedge (E_j.exp \vee E_k.exp)$
  else $E_i.exp \coloneqq E_j.exp \wedge (E_j.exp \vee E_k.exp)$
$E'_i.exp \coloneqq E_j.exp$
$E_i.pre \coloneqq \wedge.pre$
$E'_i.pre \coloneqq E_j.pre$

---

Figure 2. Attribute definition for
$\alpha_i E_i \beta_i \equiv \alpha'_i E'_i \beta'_i \rightarrow \alpha_i E_j \wedge (E_j \vee E_k) \beta_i \equiv \alpha'_i E_j \beta'_i$

With the attribute definitions, the logical expression construction algorithm is straightforward. It is simply a DFS traversal implemented as a recursion. See Algorithm 2 for details.

---

**Algorithm 2.** Expression $(V)$

Input: A syntax tree $T$ rooted at vertex $V$
Output: $V.exp$, the logical expression of $T$
1. if $V$ is a leaf and a *literal* then
2.     $V.exp = Random()$
3.     $V.pre = \infty$
4. else if $V$ is a leaf and an operator then
5.     $V.exp = V$
6.     $V.pre = $ the precedence of operator $V$
7. else
8.     for each child $U_i$ from $U_1$ to $U_n$ of $V$
9.         $Expression(U_i)$
10.    endfor
11.    Compute $V.exp$, $V.pre$ by $V$'s attribute definition
12. endif

---

Before moving to the next section, let us look at one example. Suppose the MD5 code is

$$39cf0c951da2210198e0db94f91a4b3a$$

Assume the probability $p = 0.5$. A possible production can be

$$I \Rightarrow E_0 \equiv E'_0$$

Then, the first MD5 digit is 3. We check the condition $3 < 16 \times 0.5$. So, the algorithm selects a grammar rule from $\mathcal{R}_1$. Suppose $\alpha_i E_i \beta_i \equiv \alpha'_i E'_i \beta'_i \rightarrow \alpha_i E_j \wedge E_k \beta_i \equiv \alpha'_i E_j \wedge E_k \beta'_i$ is the correct option, the next production is

$$\Rightarrow E_1 \wedge E_2 \equiv E_1 \wedge E_2$$

Suppose the third production applies Domination law on $E_1$.

$$\Rightarrow E_3 \vee T \wedge E_2 \equiv T \wedge E_2$$

On the left-hand side of $\equiv$, the disjunction is done before the conjunction. So naturally, there should be a pair of parentheses to include $(E_3 \vee T)$. However, parenthesis is not in our grammar. In fact, this is not an issue in the syntax tree. The conjunction is on the higher level.

Assume we only want to use only one equivalence law to the expression. The production starts to apply $E_i \to literal$ to end the generation.

$$\Rightarrow literal \vee T \vee E_2 \equiv T \vee E_2$$
$$\Rightarrow literal \vee T \vee literal \equiv T \vee literal$$

Once all non-terminals are substituted by terminals, a syntax tree (Figure 3) is obtained and this phase is done.

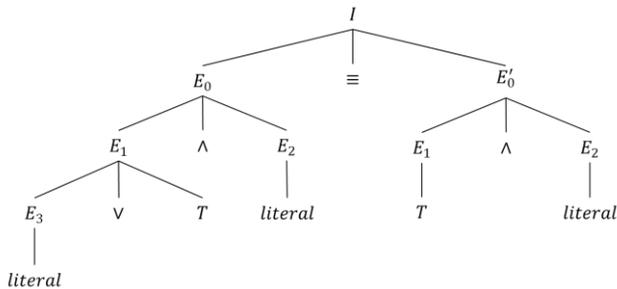

Figure 3. A syntax tree

The next phase, expression construction, converts this syntax tree into the expression $(p \vee T) \wedge p \equiv T \wedge p$ using attribute definitions.

## 3.4 Difficulty control

The syntax tree generated by phase 2 depends on the MD5 encoding of students' information. Thus, some unlucky students may receive harder questions than others. To unify the difficulties, the syntax tree generation works with a difficulty control, which concerns three following issues
- the difficulty of applying the equivalence laws,
- the total length of the entire expression, and
- the number of equivalence laws used during the production.

We notice that applying some equivalence laws is harder to students than other equivalence laws. For example, student usually believe that applying Identity $p \vee F \equiv p$ is trivial, while applying Absorption $p \vee (p \wedge q) \equiv p$ is very difficult. With this intuition, the equivalence laws are ranked into three categories as follow.

| Easy   | Identity        |
|--------|-----------------|
|        | Double Negation |
|        | Domination      |
| Median | De Morgan       |
|        | Distributive    |
|        | Idempotent      |
|        | Negation        |
| Hard   | Absorption      |
|        | Commutative     |
|        | Associative     |

Commutative and Associative are usually trivial. So, we design that every Commutative and Associative is directly followed by another equivalence law, which can be Absorption, Idempotent, Double Negation, or Identity. This technique increases the difficulty of Commutative and Associative. The number of applied equivalence laws from each category is controlled by hyperparameters from course instructors.

As for the second issue - the length of the expression, syntax tree generation terminates if and only if each non-terminal $E_i$ is replaced by a $literal$. Thus, once the length of the expression is greater than the minimum required length, we raise up the probability of choosing the production rule $E_i \to literal$ for a certain rate. The minimum expression length and the probability increment are both hyperparameters given by course instructors.

Similar to the second issue, the number of equivalence law applications is also controlled by hyperparameters. Initially, the probability of applying an equivalence law is low. In each iteration of the syntax tree generation, if an equivalence law is not selected, the probability is increased in the next iteration; otherwise, drops back to the initial value. Furthermore, if the number of applications is also limited by a maximum value.

## 4 Experiment

Our experiment aims to assess the quality and difficulty of problems generated by our algorithm and validate whether they are comparable to the human-designed problems. 40 volunteers have participated in our experiment. Each of them is assigned 3 generated questions. The answers are marked by instructors. Then, we collect historical performance over three different semesters, 2021 Spring, 2021 Fall, and 2023 Spring. Simple statistics and comparison show that the accuracy of generated questions is slightly above the minimum historical accuracy. Thus, we are confident that the generated questions can serve as good practices for students without being too frustrating. See Table 3 for details.

| Student Batch  | Participants          | Accuracy                            |
|----------------|----------------------|-------------------------------------|
| Our Experiment | 40 (120 questions)   | 61.67%                              |
| 2021 Spring    | 179                  | 68.33% (Assignment) 66.1% (Quiz)    |
| 2021 Fall      | 104                  | 80.33% (Assignment)                 |
| 2023 Spring    | 194                  | 60.98% (Assignment)                 |

Table 3 Experiment Result

## 5 Conclusion

In this paper, we provide a framework for a new automatic generation of propositional logical questions with a difficulty control based on the user input. Our system is able to generate distinct questions for different students efficiently. Experiment results provide a solid evidence for the feasibility of our generator, affirming that our approach has its potential to produce homework assignments with appropriate difficulty and quality.

In the future, we will try to design an auto grader to mark answers. Then, the entire learning evaluation process will be unmanned. From the application prospective, we also aim to develop a user-friendly front-end interface. Furthermore, we are seeking for a metalanguage which can universally describe multiple mathematics questions. Once it is done, we can design a universal question generator for other question types, even for other math courses.

## ACKNOWLEDGMENTS